\documentclass[conference]{IEEEtran}
\usepackage{times}

\usepackage[numbers,sort&compress]{natbib}
\usepackage{multicol}
\usepackage[bookmarks=true]{hyperref}
\usepackage{amsmath,amssymb}
\usepackage{graphicx}
\usepackage{booktabs}
\usepackage{algorithm}
\usepackage{algpseudocode}

\newif\ifarxivversion
\arxivversiontrue
\newif\ifincludeappendix
\includeappendixfalse

\ifarxivversion
\hypersetup{
  pdfauthor={Yina Jian},
  pdftitle={Where Do You Mean? Relational Belief Communication for Functional Robot Cooperation},
  pdfsubject={Robot decision making under uncertainty},
  pdfkeywords={spatial belief, relational communication, heterogeneous robots, decentralized POMDP, value of information}
}
\else
\hypersetup{
  pdfauthor={Anonymous Authors},
  pdftitle={Where Do You Mean? Relational Belief Communication for Functional Robot Cooperation},
  pdfsubject={Robot decision making under uncertainty},
  pdfkeywords={spatial belief, relational communication, heterogeneous robots, decentralized POMDP, value of information}
}
\fi

\begin{document}

\title{Where Do You Mean? Relational Belief Communication\\
for Functional Robot Cooperation}

\ifarxivversion
\author{%
  \IEEEauthorblockN{Yina Jian}
  \IEEEauthorblockA{Columbia University and Stanford University\\
  \texttt{yj2713@columbia.edu; yinajan@stanford.edu}}}
\else
\author{Author Names Omitted for Anonymous Review. Paper-ID [add your ID here]}
\fi

\maketitle

\begin{abstract}
Spatial intelligence in a robot team depends on whether evidence from one
viewpoint can guide another robot's actions.  In functional cooperation,
robots have different observations, capabilities, costs, and beliefs, and no
fixed cooperative sequence is supplied.  We ask whether functionally
heterogeneous foundation-model agents can construct, revise, and exploit
complementary spatial, behavioral, and action-consequence beliefs to
self-organize in unseen environments.  An overhead observer may see a target
but cannot move it, while mobile teammates can pick, push, throw, catch, or
transport, yet no single agent can finish the task.

We introduce Task-Relevant Relational Belief Communication (TRBC), a
human-inspired framework in which each agent uses partial visual observations,
ordinary dialogue, and task context to maintain private beliefs and a bounded
model of what a particular teammate may know and intend.  Messages are treated
as uncertain testimony rather than shared state.  Agents compare acting,
sharing, asking, and remaining silent by their predicted effect on a shared
cost-sensitive return.  Multimodal foundation models propose referents,
relations, utterances, and semantic options, while explicit belief updates and
a deterministic verifier retain numerical and execution authority.  The
inter-agent channel excludes coordinates, maps, structured tokens, and motor
commands.  We define a frozen protocol spanning pick--load--deliver,
obstacle-clearing, physical toss--relay, and held-out task compositions under
procedural layouts, occlusion, physical failures, and matched resource budgets.
It tests whether distributed belief supports grounded dialogue, adaptive
division of labor, recovery, and delivery rather than fluent execution of a
fixed script.
\end{abstract}

\ifarxivversion
\begin{center}
\small\emph{Work in progress. This version presents the problem formulation,
TRBC architecture, implemented physical-system core, and frozen evaluation
protocol. Multi-skill integration and full-scale experiments are ongoing.}
\end{center}
\fi

\IEEEpeerreviewmaketitle

\section{Introduction}
An overhead observer sees a target behind a tree but cannot pick it.  A mobile
picker can retrieve the target, yet its ground-level camera may see neither the
target nor the tree.  A separate carrier can transport a loaded basket to the
goal, but it cannot grasp the target or reliably infer whether loading
succeeded. The observer can say that the target is left of the red tree and
beyond the green rock. The picker must ground that description in its own view
and decide where to search. Later, if the target slips before loading, the
carrier must use the picker's report to decide whether to wait or depart.
At both points, one agent holds evidence another needs for its next action.

Now place a movable rock across the bridge and give the team two possible
receivers: a passive basket cart and a carrier that can extend a physical net.
The team might spend fuel to bring a pusher to clear the bridge, or a thrower
might relay the target to the catcher if the agents can agree on a landmark-
relative rendezvous and readiness signal.  Neither solution is prescribed.
Which one is sensible depends on private views, remaining fuel, action costs,
object affordances, and uncertain push, throw, and catch outcomes.  A long
horizon here is therefore not a fixed chain of roles but an execution trace
that emerges from dialogue, belief, and physical feedback.

This example exposes a general systems problem at the intersection of spatial
intelligence and robot cooperation.  Functionally different robots do not need
the same state estimate.  The observer needs target--landmark relations; the
picker or thrower needs a local target and reachable manipulation; the pusher
needs obstacle affordance and clearance consequences; and a cart or catcher
needs rendezvous, readiness, load status, and a traversable route.  No agent
observes the complete task state, and no agent can finish alone.  A map can
support navigation, but deciding whether to clear, carry, throw, catch, wait,
or ask also depends on task progress, teammate behavior, cost, and execution
outcomes.  Each role needs a belief suited to its own decision and a way to
obtain relevant evidence from its teammates.

\begin{figure}[!t]
  \centering
  \includegraphics[width=\columnwidth]{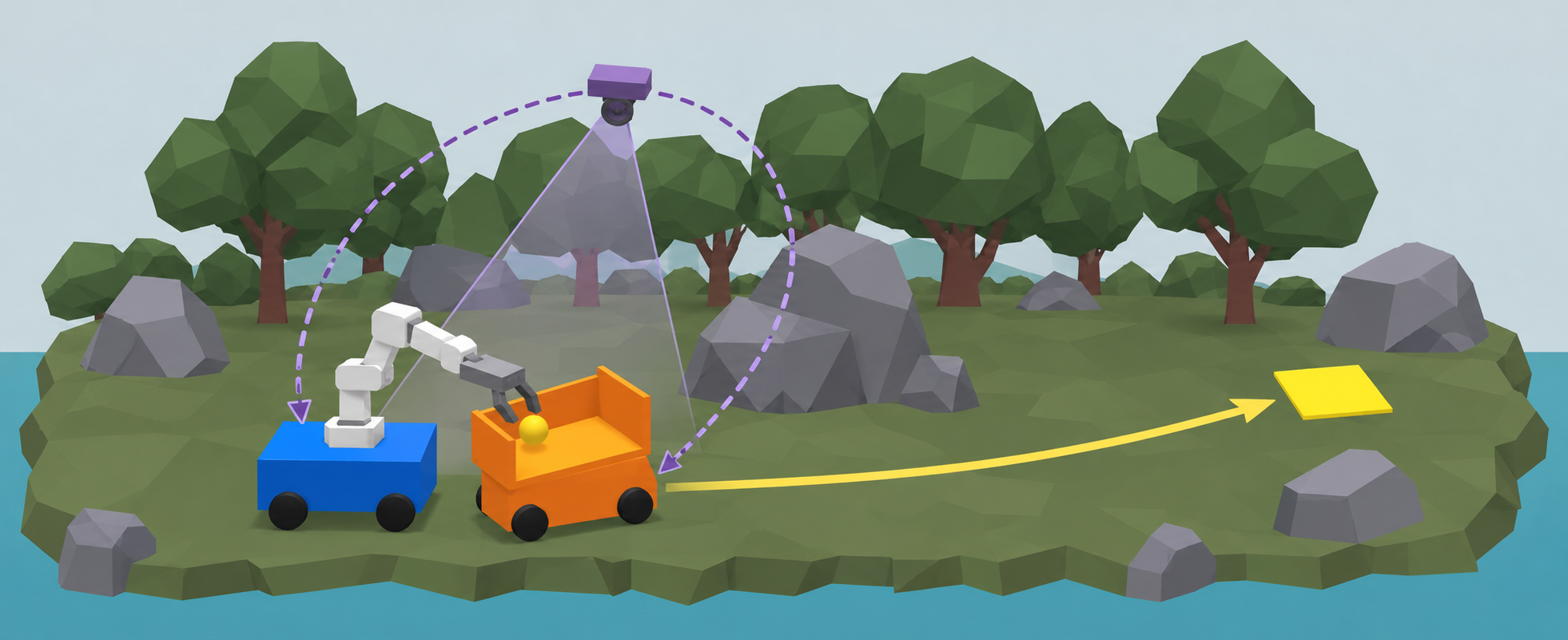}\par
  \smallskip
  \noindent\hbox to \columnwidth{%
    \shortstack{\includegraphics[height=0.205\columnwidth]{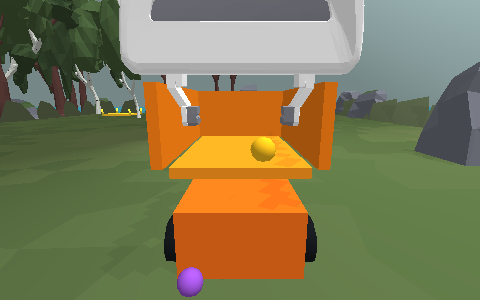}\\\footnotesize Picker}\hfil
    \shortstack{\includegraphics[height=0.205\columnwidth]{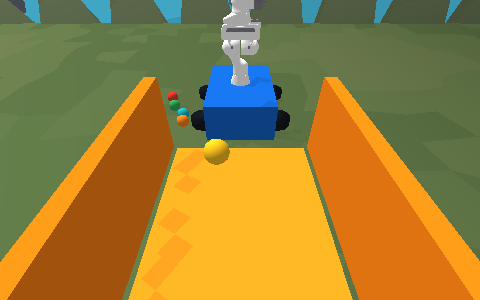}\\\footnotesize Carrier}\hfil
    \shortstack{\includegraphics[height=0.205\columnwidth]{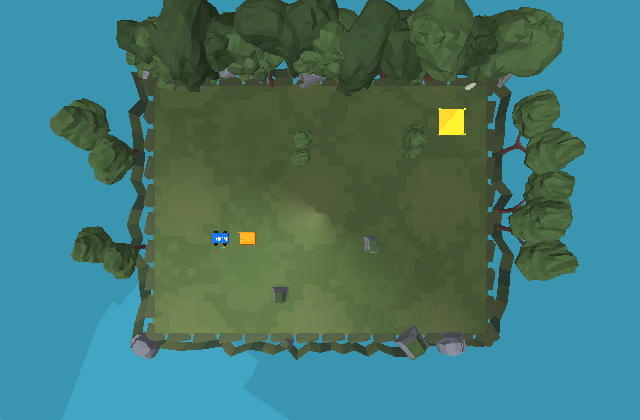}\\\footnotesize Observer}%
  }\par
  \caption{\textbf{Complementary spatial beliefs for functional cooperation.}
  \textbf{Top:} A conceptual illustration of the overhead observer (purple),
  mobile picker (blue), and basket carrier (orange). Dashed links represent
  relational evidence exchanged between roles.
  \textbf{Bottom:} Synchronized PyBullet RGB frames immediately after the
  first load in the scripted functionality run (seed 11). Left to right:
  picker forward view, carrier basket inspection view, and observer overhead
  view. The two local views show the same loaded particle from opposite sides.}
  \label{fig:system-overview}
\end{figure}

\emph{Theory of Space} studies how an agent constructs, revises, and exploits a
spatial belief from successive partial views during active exploration, and
identifies belief drift and inertia in foundation models
\citep{zhang2026theoryofspace}. \emph{Seeing What Execution Needs} asks which
visual evidence makes a given manipulation command executable
\citep{jian2026seeing}. Functional cooperation distributes this cycle across
roles. One agent may hold the evidence another needs, and its relevance
depends on the recipient's capabilities and current subtask.

We therefore ask:
\begin{quote}
\emph{Can functionally heterogeneous foundation-model agents construct,
revise, and exploit complementary beliefs from incomplete visual evidence,
ordinary dialogue, and physical feedback to self-organize cost-sensitive,
long-horizon cooperation in unseen environments?}
\end{quote}
We call the required representation \emph{distributed actionable spatial
belief}: complementary private beliefs whose quality is measured by whether
they support appropriate role-specific actions. A target--landmark relation
can guide a picker or thrower during search.  A readiness report can determine
when a catch is attempted; a load-status report can determine whether a carrier
waits or departs; and a failed push can change whether the team clears a bridge
or selects a relay.  This makes actionability depend on what the recipient
believes, can do, and needs to decide next.  Each agent must decide whether to
act, seek evidence, share evidence, or remain silent on that basis.

We study this question with procedurally composed teams in randomized PyBullet
workspaces.  Each episode assigns finite skill inventories---observation,
pick--place, pushing, throwing, cart transport, or net-assisted catching---and
samples a scene in which no single inventory is sufficient.  Only a delivery-
capable carrier scores by bringing the correct target into the goal.
Our approach is \emph{Task-Relevant Relational Belief Communication} (TRBC).
Each agent maintains a private relational scene graph, spatial, behavioral,
and action-consequence beliefs, and a bounded first-order model of what a
specific teammate may know.
Each
role uses TRBC to compare physical actions, addressed messages, clarification
queries, and silence on a common expected-team-return scale.

Our contributions are:
\begin{itemize}
  \item a formulation of \emph{distributed actionable spatial belief} that
  extends task-conditioned spatial reasoning to long-horizon cooperation among
  agents with complementary observations, capabilities, and evidence needs;
  \item TRBC, a belief-mediated architecture for calibrated cross-view
  grounding, human-compatible dialogue, belief revision, recipient-conditioned
  communication, clarification, and verified physical execution; and
  \item an interactive multi-task evaluation protocol that links belief quality
  to emergent division of labor, physical handoff, delivery, recovery, and
  resource efficiency across unseen layouts and capability compositions.
\end{itemize}

Figure~\ref{fig:trbc-overview} makes the experimental scope concrete.  The
three columns instantiate different capability compositions and physical
handoffs, while the shared row at the bottom emphasizes that the coordination
mechanism is unchanged.  The numbered panels are illustrative outcomes of
role-local decisions rather than a task graph supplied to the agents.  This
extends the construct--revise--exploit view of spatial belief
\citep{zhang2026theoryofspace} and task-conditioned actionability
\citep{jian2026seeing} from one embodied reasoner to a team whose evidence,
skills, and consequences are distributed across roles.

\section{Related Work}
\paragraph{From spatial representation to actionable belief.}
Spatial action and intention maps align state, action, and communicated intent
in a common overhead image \citep{wu2021spatialintention}.  Recent foundation-
model work instead asks whether an agent can form a spatial model from limited
views: RoboSpatial studies relational grounding \citep{song2025robospatial},
MindCube probes spatial mental models \citep{yin2025mindcube}, and \emph{Theory
of Space} decomposes belief use into construction, revision, and exploitation
\citep{zhang2026theoryofspace}.  ESI-Bench and BEHAVIOR-Spatial connect spatial
reasoning to embodied behavior \citep{hong2026esibench,
chen2026behaviorspatial}; active-perception systems learn which view improves
manipulation \citep{xiong2025vision,liu2026activevla,bai2026taskaware}.
Task-conditioned actionable belief further judges evidence by whether it makes
a downstream skill executable \citep{jian2026seeing}.  These formulations
either provide a shared spatial canvas or study one decision-maker.  We study
the complementary case: useful evidence originates in one private view, must
be grounded in another, and is actionable only relative to the recipient's
physical role.

\paragraph{Partial observation and private world models.}
PARTNR exposes planning and recovery failures in heterogeneous embodied teams
\citep{chang2025partnr}, while COMBO reconstructs task-relevant global state
from partial egocentric observations \citep{zhang2025combo}.  Seeing Together
fuses synchronized egocentric video for cooperative spatial question answering
\citep{peng2026seeingtogether}.  More directly, dialogue has been used to align
agents' private world graphs and to measure observation convergence,
information novelty, and belief-sensitive messaging
\citep{dongre2026aligning}.  Its finding that dialogue can reduce action
conflicts while degrading task success motivates explicit belief updates
rather than treating conversational agreement as alignment.  Controlled PARTNR
studies separately vary partner observation, object-knowledge exchange, and
next-action prediction \citep{szlatincsan2026observe}.  These works reveal that
successful dialogue need not imply aligned internal models.  Our objective is
not full world-model consensus: each role retains a different probabilistic
relational belief, and communication is evaluated by whether it changes the
correct role-specific physical decision.  Contact outcomes such as a missed
grasp, drop, or empty basket then revise spatial, behavioral, and
action-consequence hypotheses.  Seeing Isn't Believing identifies the
complementary failure of belief inertia and mitigates it through
estimate--verify--update interventions \citep{wang2026seeing}; it operates on a
single embodied agent's textual belief, whereas we study revision across
agents whose visual evidence and physical feedback disagree.

\paragraph{Semantic coordination among heterogeneous robots.}
RoCo uses dialectic LLM discussion for multi-robot planning
\citep{mandi2024roco}, and EMOS conditions planning on embodiment and capability
\citep{chen2025emos}.  Recent systems learn natural-language coordination from
multimodal demonstrations \citep{goldfajn2026robotalk}, combine distributed VLM
orchestration with VLA execution \citep{zhou2026duomind}, or use need-driven
dialogue to negotiate task allocation and team composition
\citep{li2026proactive}.  HEROIC demonstrates natural-language-only coordination
without a shared map, using evidential search beliefs in heterogeneous
air--ground teams \citep{chauhan2026heroic}.  These systems establish that
semantic communication can allocate skills, synchronize plans, and support
search.  TRBC addresses a different bottleneck.  DuoMind's RoboPoly agents also
receive a common global camera view, while HEROIC targets open-vocabulary
identification rather than a manipulation handoff.  In TRBC, no message, plan,
or global view is accepted as shared state.  Observer, picker, and carrier
independently interpret ordinary language against private visual evidence, and
execution feedback can overturn what fluent dialogue previously suggested.

\paragraph{Decision- and belief-aware communication.}
Value-of-information policies ask when communication improves decentralized
POMDP return enough to justify its cost \citep{trombetta2026asking}, while
learned cooperative policies select task-relevant state to transmit
\citep{bae2026llmcommunication}.  LatentToM learns ego and consensus
representations for cooperative manipulation \citep{he2025latenttom}.  CRAFT
tests pragmatic multi-sender communication when each speaker observes only one
projection of a construction target \citep{nath2026craft}, but its discrete
block construction lacks contact dynamics and execution-conditioned belief
revision.  CommCP calibrates LLM messages for multi-robot embodied question
answering \citep{zhang2026commcp}.  Thus theory of mind, communication value,
and message calibration are not individually new.  Our contribution is their
coupling to cross-view relational evidence in closed-loop physical cooperation:
the same utterance is valued differently for a picker and carrier, enters the
recipient's belief as uncertain evidence rather than copied state, and is
tested against subsequent action and physical outcome.

\paragraph{Position of this work.}
The open gap is not another conversational planner, nor a language interface
to a shared map.  It is whether functionally interdependent agents can use
human-compatible dialogue to turn different, incomplete visual histories and
testimony into private spatial, behavioral, and causal beliefs that remain
useful after the world pushes back, then organize a joint strategy without a
prescribed plan.  This distinction matters because task success can hide
shortcut coordination---for example, privileged coordinates, stable object
identifiers, or a leader's plan---without demonstrating spatial understanding.
Our held-out layouts, identifier reminting, ambiguous landmarks, belief-swap
tests, communication budgets, and physical failure interventions therefore
evaluate the mechanism itself: whether agents construct, revise, and exploit
distributed actionable belief, rather than merely agree on a script.

\section{Problem Formulation}
We study \emph{distributed actionable spatial belief}: functionally different
agents must turn partial visual observations and natural-language evidence into
complementary private beliefs that support the next role-specific decision.
Each agent maintains the evidence its current subtask requires and estimates
what a particular teammate may already know, intend, or be ready to do.  These
beliefs support local decisions that contribute to a shared goal without
prescribing a global plan.  We use \emph{human-inspired} to denote this
coordination pattern---private and potentially inconsistent beliefs repaired
through ordinary dialogue and feedback---not equivalence to human cognition.

\subsection{Functional-Cooperation Task}
Let $\mathcal R=\{O,P,U,T,C_B,C_N\}$ denote an overhead RGB observer, a mobile
picker, a pusher, a thrower, a basket-cart carrier, and a net-catcher carrier.
An episode samples a complementary team $\mathcal I\subseteq\mathcal R$ and a
finite skill inventory $\mathcal A_i$ for each member.  The observer can look,
describe, and answer but cannot manipulate.  The picker can approach, pick, and
place.  The pusher can move particles and physically movable obstacles.  The
thrower can pick and release a target toward a qualitative region.  The two
carriers share a mobile base and terminal delivery authority: $C_B$ presents
and inspects a passive basket, whereas $C_N$ orients and extends a physical net,
tracks an incoming target, retracts, and inspects.  Only a delivery-capable
carrier earns terminal reward by bringing the correct captured target into the
goal; no sampled inventory can complete the episode alone.

Each episode procedurally samples particles $X$ with a visually
distinguishable target $x^\star$, delivery region $G$, bridges or gaps,
obstacles, landmarks $L$, robot starts, fuel budgets, and physical parameters.
Landmark appearance is deliberately language-groundable: category and salient
attributes yield descriptions such as a red tree or a green rock.  Attribute,
position, affordance, and task relevance are randomized independently, so
appearance may identify a landmark but cannot encode a coordinate, route,
movability label, or policy.  Mass, friction, and restitution vary across
objects, making push, throw, and catch outcomes uncertain even for familiar
appearance.

The team receives a shared goal and cost model, not a task graph or role
sequence.  Depending on its beliefs, it may load a nearby cart, spend fuel to
clear a bridge, push a particle into reach, or coordinate a toss to an extended
net.  Search, rendezvous, manipulation, verification, transport, recovery, and
delivery are evaluator milestones, not policy inputs.  Local views may omit a
target, landmark, obstacle state, teammate readiness, or handoff outcome;
repeated attributes create referential ambiguity; and grasps, pushes, throws,
catches, and placements can fail.  At each decision event an agent may observe,
act, converse, request clarification, or wait.  Section~\ref{sec:evaluation}
describes the task families and held-out procedural generator.

\subsection{Distributed Decision Problem}
We model the task as a finite-horizon communication-enabled Dec-POMDP with
sampled agents $\mathcal I\subseteq\mathcal R$, latent physical state,
role-specific observations and skill inventories, communication actions,
shared reward, and a finite horizon.  The latent state contains target identity
and object, receiver, landmark, obstacle, bridge, and robot configurations
together with grasp, push, release, flight, catch, load, route, readiness, fuel,
and handoff status.  The overhead observation is RGB rather than privileged
state; local observations contain egocentric RGB, proprioception, remaining
private resources, and role-specific execution feedback.

Agent $i$ acts from its private observation--message--action history.  Its
internal belief state separates spatial uncertainty, teammate and handoff
status, and a bounded first-order model of what each recipient may know:
\begin{equation}
\mathcal B_t^i = \left(b_{t,\mathrm{sp}}^i,\; b_{t,\mathrm{beh}}^i,\;
b_{t,\mathrm{cau}}^i,\; \{\mu_t^{i\rightarrow j}\}_{j\neq i}\right).
\label{eq:private-belief}
\end{equation}
The spatial component is multimodal over the target, landmarks, basket, goal,
and qualitative relations.  The behavioral component represents states such
as approaching, ready, loading, departing, waiting, or failed.  The
action-consequence belief $b_{t,\mathrm{cau}}^i$ stores context-conditioned
probabilities of skill outcomes---for example grasp, release, load, and handoff
success---and is revised by attempted actions and their physical feedback.
``Causal'' is used in this operational, intervention-conditioned sense; we do
not claim unrestricted structural causal discovery from passive observations.
The recipient model is intentionally first order and does not recursively
represent what $j$ believes about $i$.

Physical options are drawn from the acting agent's sampled inventory: sensing
and description for the observer; approach, pick, and place for the picker;
particle or obstacle pushing for the pusher; pick, aim, and physical release for
the thrower; presentation, inspection, transport, and delivery for the cart;
and orientation, net extension, tracking, catch inspection, transport, and
delivery for the catcher.  Every role may additionally describe, report
uncertainty or intent, request, clarify, confirm readiness, correct a prior
claim, or remain silent.  Reward is issued only for delivering the correct
target and is reduced by movement and fuel, skill execution, elapsed time,
failure, communication, clarification, and model calls.  All resources are
budgeted, so an agent that can act is not necessarily the cheapest teammate to
act.

\subsection{Recipient-Relevant Evidence and Controlled Communication}
Let $e$ be evidence available to sender $i$, and let $\tau_j(b^j,e)$ denote
recipient $j$'s resulting belief update.  Evidence is actionable for $j$ only
when it improves the recipient's predicted team decision:
\begin{equation}
\Delta_j(e\mid b^j)=
\max_a Q_{\mathrm{team}}^j\!\left(\tau_j(b^j,e),a\right)
-\max_a Q_{\mathrm{team}}^j(b^j,a).
\label{eq:recipient-relevance}
\end{equation}
The same target--landmark relation may therefore be valuable to the picker but
irrelevant to the carrier, which may instead require basket occupancy or route
status.  Recipient relevance is defined by predicted decision change, not by
message length or entropy reduction alone.

To test spatial understanding rather than privileged state transfer, the main
condition disallows communicated coordinates, maps, waypoints, trajectories,
serialized records, and motor commands.  Low-level controllers may use private
metric geometry, but it cannot enter model prompts or messages.  The channel
carries only ordinary natural-language observations, uncertainty, intentions,
requests, readiness confirmations, outcome reports, and corrections.  For
example, a thrower may say, ``I will release after your net is open,'' and the
catcher may answer, ``I am beside the red rock, but I cannot yet see the
target.''  Recipients ground the original utterances in their own observations
and may preserve multiple hypotheses, disagree, or report that a referent is
unknown.  Private parsers may create internal typed evidence after reception,
but agents cannot exchange JSON, posterior values, evidence IDs, hidden skill
state, code-like strings, or other machine-only channels.  Coordinate and
shared-map communication are retained only as oracle controls.  This is a
human-compatible dialogue interface, not a claim that the agents reproduce
human cognition.

\subsection{Research Hypotheses}
Evaluation follows the construct--revise--exploit decomposition.
\textbf{H1---Construct:} explicit relational spatial belief improves target
grounding, rendezvous, and delivery on held-out layouts and capability
compositions over history-only and free-form agents.  \textbf{H2---Revise:}
calibrated message and execution updates reduce stale-belief failures,
including wrong-target approaches, repeated failed grasps or pushes, missed
throws or catches, incorrect placement, and empty-carrier departure.
\textbf{H3---Exploit:} recipient-conditioned communication and clarification
achieve a better task-return--communication-cost frontier than broadcast,
always-ask, and entropy-triggered policies under matched budgets.
\textbf{H4---Reorganize:} under changes to starts, fuel, obstacle affordance,
receiver embodiment, or physical outcome, the same frozen TRBC policy selects
different valid divisions of labor and recovers more often than a fixed skill
chain.

\begin{figure*}[!t]
  \centering
  \includegraphics[width=0.98\textwidth]{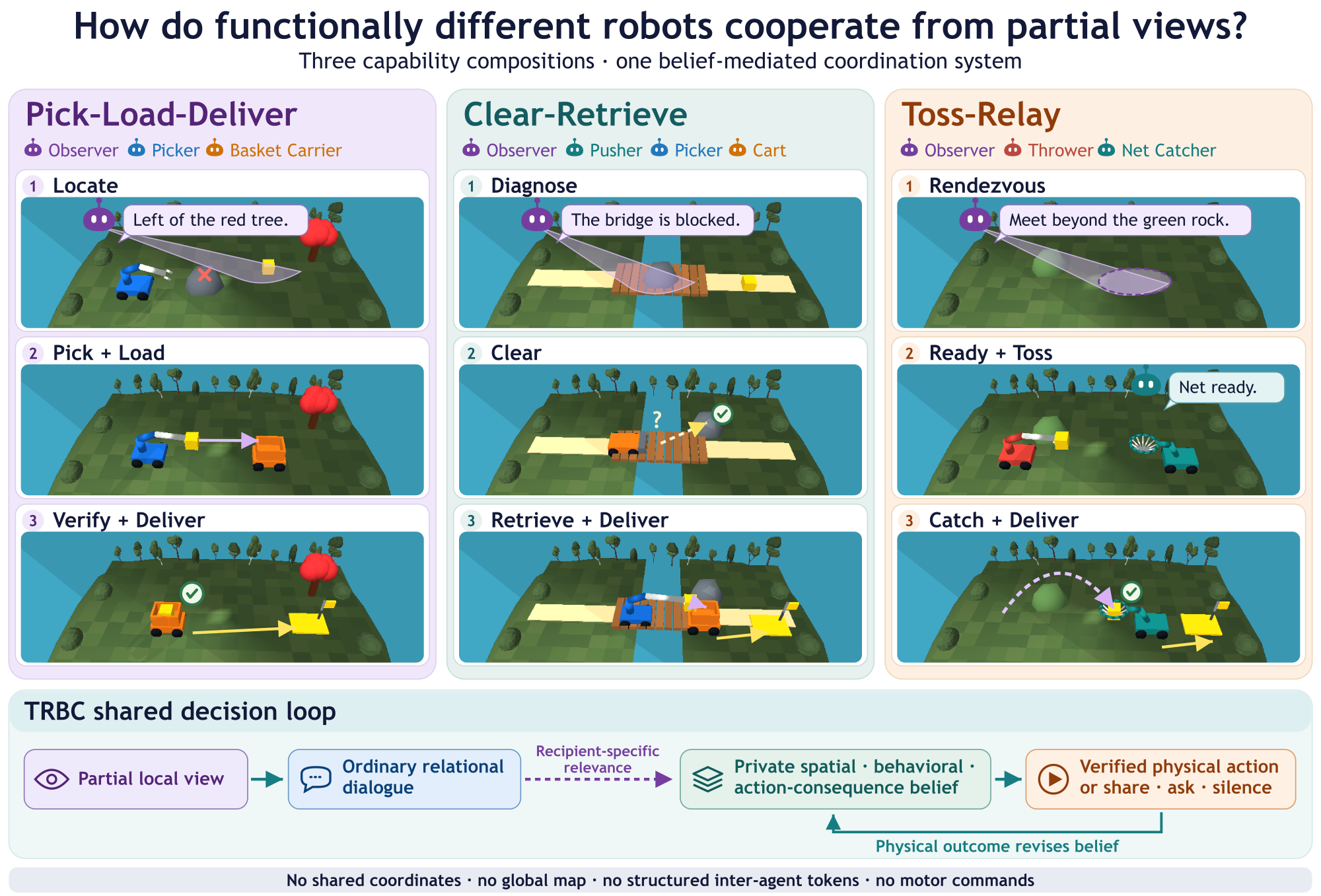}
  \caption{\textbf{Three functional compositions governed by one
  belief-mediated coordination system.}  Pick--load--deliver combines an
  overhead observer, mobile picker, and basket carrier.  Clear--retrieve adds
  a pusher whose intervention changes route feasibility.  Toss--relay pairs a
  thrower with a net catcher at a language-grounded rendezvous.  Each column
  shows one possible successful episode, not a prescribed skill sequence.
  Across all compositions, agents receive partial local views, exchange only
  addressed relational dialogue, maintain recipient-specific private spatial,
  behavioral, and action-consequence beliefs, and revise those beliefs after
  physical execution.  No shared coordinates, global map, structured
  inter-agent tokens, or motor commands are communicated.}
  \label{fig:trbc-overview}
\end{figure*}

\section{Method: Task-Relevant Relational Belief Communication}
TRBC helps a robot decide what to do when a teammate holds the evidence it
needs (Fig.~\ref{fig:trbc-overview}, bottom). If the picker observes a drop, its report
can lower the carrier's belief
that the basket contains the target and make checking preferable to departing.
During search, the same mechanism lets a manipulator ground an overhead
description against local landmarks and request clarification when plausible
interpretations lead to different approaches.  During handoff, it lets a
thrower wait for a catcher's readiness or lets the team abandon a costly
bridge-clearing attempt.  Each role maintains its own belief, revises it with
local and communicated evidence, and chooses from its own skills and dialogue
actions.  No module supplies a global task sequence or central motor plan.
Foundation models propose interpretations, human-readable utterances, and
semantic options; explicit updates, cost-sensitive candidate scoring, and
execution checks determine the decision.

\subsection{Role-Specific Belief and Memory}
Each agent stores a private relational scene graph, the belief components
in Eq.~\eqref{eq:private-belief}, an evidence ledger, and remaining action,
movement/fuel, message, query, and model-call budgets, together with its own
skill inventory and a bounded model of teammate capability and readiness.
Scene-graph nodes represent perceived objects, appearance, visibility, and
uncertainty; edges carry distributions over a finite qualitative relation
vocabulary and reference frames.  Similar landmarks remain separate
association hypotheses until evidence resolves them.
The evidence ledger records which local observations, utterances, and physical
outcomes have been consumed so duplicates and paraphrases are not counted
again.

A picker can tell the carrier, \emph{I saw the target fall before loading.}
The carrier interprets this report as evidence about its own load status and
records that interpretation in a private \texttt{SpatialToken}. The runtime
attaches the delivery event and any locally grounded support. The sender's
referent IDs, evidence records, and belief values stay in its own runtime.
The internal \texttt{trbc.v1} records cover role memory, interpreted evidence,
proposals, verifier decisions, and trajectory events. Only addressed natural
language reaches the recipient. Internally, the implementation logs role
memory, recipient-grounded evidence, candidate proposals, verifier decisions,
and trajectory events; these structured records remain private and never cross
the inter-agent channel.

\subsection{Calibrated Cross-View Belief Revision}
A recipient grounds each utterance in its own view, treats it as calibrated
evidence, and keeps multiple hypotheses when the reference is ambiguous.
Physical outcomes provide new evidence for later
decisions.

At each decision event, an agent first predicts from its previous posterior and
incorporates one new local observation.  It then grounds each new utterance.
Let $L_m^j(x)$ be recipient $j$'s spatial likelihood after marginalizing
candidate landmark associations and reference frames for utterance $m$.
Language is uncertain evidence rather than fact, so the message posterior is
updated with
\begin{equation}
b_{t,+}^j(x) \propto b_{t,\mathrm{local}}^j(x)
\left[\kappa_m L_m^j(x)+(1-\kappa_m)U^j(x)\right]^{\alpha_m},
\label{eq:calibrated-message}
\end{equation}
where $\kappa_m$ is a calibrated reliability for the sender and message family,
$U^j$ is a feasible-region null likelihood, and $\alpha_m$ discounts dependent
evidence.  Raw model confidence is never treated as probability.  Unparseable
language has no belief effect, ambiguous language preserves multiple modes,
and expired or already-consumed evidence is rejected.  Execution feedback
updates only the affected spatial, behavioral, or action-consequence
predicates; a missed grasp, for example, need not erase an unrelated landmark
relation. Updates follow a single-consumption sequence---prediction, one local
observation, each unseen message once, and physical feedback---with evidence
identifiers, expiration, sender calibration, association and reference-frame
marginalization, and dependence discounting preventing duplicate or
overconfident updates.

\begin{figure}[!t]
  \centering
  \includegraphics[width=0.98\columnwidth]{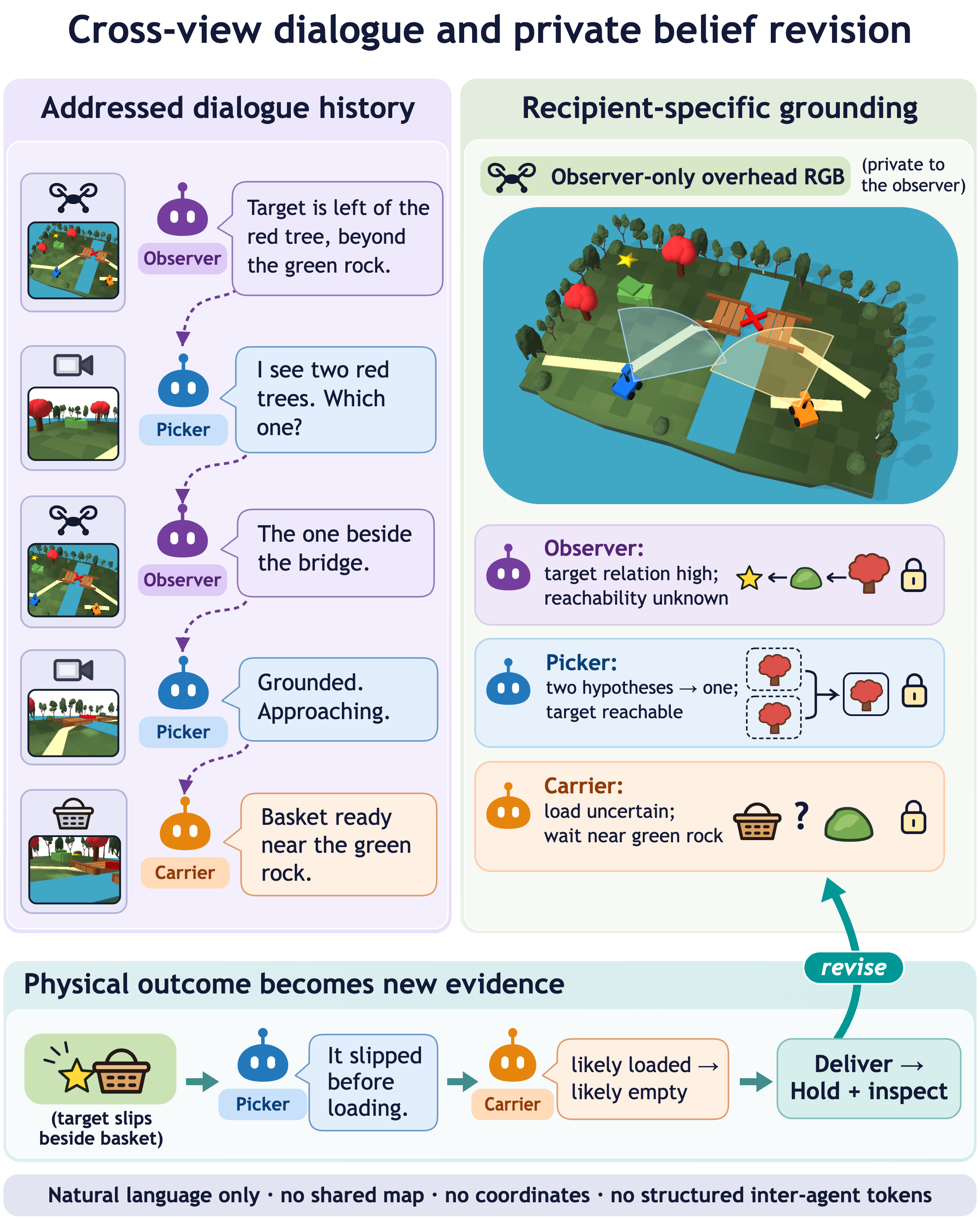}
  \caption{\textbf{Cross-view dialogue and recipient-private belief
  revision.}  Left: ordinary addressed dialogue resolves an ambiguous
  landmark description across observer and picker views, then lets the carrier
  report its own readiness.  Right: the observer, picker, and carrier retain
  different grounded hypotheses because their observations, capabilities, and
  pending decisions differ.  Bottom: a target slip becomes new evidence.  The
  picker's natural-language report lowers the carrier's load belief and changes
  its preferred action from delivery to holding and inspection.  Camera tiles
  are direct PyBullet RGB renderings.  Role icons, speech bubbles, private-
  belief diagrams, and translucent fields of view are explanatory overlays and
  are not policy observations.}
  \label{fig:dialogue-belief-revision}
\end{figure}

Figure~\ref{fig:dialogue-belief-revision} visualizes how cross-view dialogue
becomes recipient-private evidence and is revised by physical execution.

\subsection{Recipient-Conditioned Action and Communication}
Each role generates bounded candidates from four classes: physical options,
addressed messages, clarification queries, and silence.  Physical candidates
must satisfy role capability and belief preconditions.  Messages must be
supported by unconsumed evidence.  Queries target a relation, landmark
association, reference frame, or behavioral predicate whose resolution could
change the next physical choice.  For every candidate, TRBC estimates team
return under the acting agent's private belief and recipient model, then makes
one unified decision:
\begin{equation}
c_t^i=\arg\max_{c\in\mathcal C_t^i}
\left[\widehat Q_{\mathrm{team}}^i(c)-C(c)\right].
\label{eq:unified-selection}
\end{equation}
For a message, value of communication is its estimated score minus the score of
silence for the same recipient.  For a query, value of information is its
expected post-reply score minus the best current physical score.  These are
reported for interpretation but are not separate objectives.  Communication
or clarification is selected only when it improves the same return used to
rank physical action. The bounded approximation uses recipient-belief
particles and one-step macro-action lookahead; messages are compared with
silence and queries with the best currently available physical action under
identical resource costs.

\subsection{Foundation-Model Proposals and Verification}
A role-conditioned multimodal model proposes referents, relations, utterance
intents, and semantic options. Its input is a bounded private memory summary,
its role's observations, addressed dialogue, the public goal, and task context
inferred from its own history, together with its own skills, remaining resource
bins, and the allowed relation and action vocabulary.  It receives neither a
task graph nor a recommended global sequence.  Posterior values, evidence IDs,
utility scores, and controller bindings stay in the runtime.

\textbf{Observation and execution boundary.} The observer sees top-down RGB.
Mobile agents receive egocentric RGB, proprioception, private resource status,
and their own sensor-derived skill feedback, such as gripper contact, push
resistance, release, net contact, or visible load cues.  RGB comes without
simulator overlays or segmentation IDs. Exact poses of unobserved objects or
teammates, exact basket/net occupancy, and hidden task milestones are
evaluator-side state. A role chooses a semantic option, such as approaching a
grounded landmark, pushing a perceived obstacle, extending a net, or holding
position. Its low-level skill may use metric geometry derived from local
sensing and odometry and returns only role-observable feedback. The simulator
resolves physical contacts. An isolated evaluator checks success from ground
truth. Only the labeled oracle conditions provide privileged state to policies.

\textbf{Proposal and communication checks.} A closed proposal schema limits
outputs to known action and relation types. The runtime checks references
against the sender's available evidence, but the transmitted payload is the
model's ordinary sentence rather than the proposal schema.  Agents may express
observation, uncertainty, intention, readiness, requests, failure, and
correction using a bounded qualitative relation vocabulary and role-local
landmark descriptions.  Addressing is routing metadata. Message payloads carry
no coordinates, waypoints, serialized records, posterior values, hidden IDs,
or motor commands. Execution
checks enforce role capability, valid skill binding, observable feasibility,
and budgets. A separate TRBC check filters unsupported, stale, and consumed
evidence before belief updates. Verified model options join the deterministic
candidates and are scored by Eq.~\eqref{eq:unified-selection}. Invalid outputs
and provider failures leave deterministic options available. This interface
restricts information flow. E4 tests shortcut use within allowed inputs.

\subsection{Online TRBC Loop}
Algorithm~\ref{alg:trbc} presents the main system loop.  Each observation and
evidence item is incorporated at most once; the physical outcome becomes
explicit evidence for the next belief revision.
\begin{algorithm}[t]
\caption{Task-Relevant Relational Belief Communication}
\label{alg:trbc}
\begin{algorithmic}[1]
\Require role $i$, private memory $M_0^i$, skills, budgets, shared goal
\While{task is active and budget remains}
  \State Observe locally and update the private scene graph
  \State Update spatial, behavioral, and action-consequence beliefs once
  from local evidence
  \State Ground new utterances; reject unsupported, stale, or duplicate evidence
  \State Apply calibrated message updates and revise recipient models
  \State Generate physical, message, query, and silence candidates
  \State Add role-conditioned foundation-model semantic proposals
  \State Verify capabilities, evidence, preconditions, and budgets
  \State Select $c_t^i$ with Eq.~\eqref{eq:unified-selection} and execute
  \State Update belief from the physical outcome and append a trajectory event
\EndWhile
\end{algorithmic}
\end{algorithm}

\subsection{Assumptions and Computational Cost}
The main setting assumes cooperative agents, shared task reward, known role
capabilities, a finite relation vocabulary, and static landmarks within an
episode.  Delivery is reliable but content may be ambiguous, stale, or wrong;
packet loss and adversarial senders are stress tests.  Recipient reasoning uses
a bounded first-order particle approximation, and action values use one-step
macro-action lookahead.  Candidate caps, particle count, expiration horizons,
calibration maps, prompts, and budgets are frozen on development layouts.
Grounding cost is linear in the bounded message count and retained graph
relations; candidate valuation is linear in candidate count and recipient
particles. The frozen verifier rejects capability mismatches; unsupported,
stale, or duplicate evidence; invalid physical preconditions; coordinate or
map leakage; structured inter-agent payloads; and budget violations. All
accepted and rejected candidates and resulting trajectory events are logged.

\section{Experimental Design and Evaluation Protocol}\label{sec:evaluation}
Our evaluation follows the construct--revise--exploit decomposition. First,
we test whether explicit beliefs improve target grounding and rendezvous in
held-out layouts and team compositions.  Second, we introduce contradictory
visual and physical evidence to test whether beliefs and subsequent actions
change after wrong-target approaches, failed grasps or pushes, missed throws
or catches, drops, and empty-carrier handoffs.
Third, under matched message and model-call budgets, we test whether
recipient-conditioned communication improves the task-return--communication-
cost frontier over broadcast, always-ask, entropy-triggered, history-only, and
free-form dialogue policies.  Fourth, paired interventions on starts, fuel,
receiver embodiment, obstacle affordance, and execution outcomes test whether
the same policy reorganizes division of labor rather than replaying a fixed
skill chain.  Counterfactual belief swaps, landmark-attribute permutations,
and coordinate and centralized-state oracles distinguish belief-mediated
reasoning from visual or linguistic shortcut mappings.

We freeze the environment generator, prompt templates, relation vocabulary,
belief-update parameters, and communication budgets before evaluating the
primary hypotheses.  Non-oracle methods share the role-specific observation interface, low-level
skills, execution checks, horizon, and matched scene seeds.  The same TRBC
prompt and update code is used across task families; only the public goal,
private role inventory, embodiment, and observations change.  Views can diverge
after different actions. Paired decision replays hold the view and available
evidence fixed at the decision point.

\subsection{Procedural PyBullet Environment}
An RGB-only overhead observer supplies broad visual context but cannot
manipulate.  Mobile embodiments draw from pick--place, particle/obstacle push,
physical throw, basket-cart receive, net-catch receive, transport, and delivery
skills.  The two carrier variants share a chassis and terminal scoring rule but
change the handoff mechanism: the cart accepts contact placement into a passive
basket, whereas the catcher must orient, extend a collision-enabled net, catch,
retract, inspect, and then transport.  Agents communicate through ordinary
natural language.  The main condition excludes coordinates, maps, structured
records, trajectories, and motor commands; each role grounds what it hears in
its own view and autonomously chooses its next physical or dialogue action.

Every episode samples island geometry, bridges and gaps, obstacle density,
landmark categories and attributes, target and goal poses, receiver embodiment,
team skill allocation, robot starts, fuel budgets, camera occlusions, and
physical parameters including mass, friction, and restitution.  The
blower/pusher disturbance phase runs before the first cooperative observation
and is replayed identically for all methods on a matched seed.  Development and
test splits differ not only in seed but also in layout topology, landmark
composition, affordance combinations, and held-out capability composition.
Landmark color, position, function, and traversability are independently
permuted, preventing appearance from becoming a location or policy code.

Push, throw, and catch use simulator contacts rather than attachment or
teleportation.  A pusher must contact the particle or obstacle; a thrower must
physically release; and a catch requires contact with the deployed net under
admissible position and velocity.  Targets may bounce, miss, or fall out, and a
carrier must inspect before deciding to depart.  A privileged scripted oracle
first verifies that each sampled scene is physically solvable, but its state
and plan remain evaluator-only.

\subsection{Task and Capability Suite}
We use one method and four primary task families.  \textbf{Pick--Load--Deliver}
pairs a picker with the basket cart.  \textbf{Clear--Retrieve} adds a pusher
that may move a particle or clear a movable rock from a bridge; object mass,
friction, and push direction determine whether clearance succeeds.
\textbf{Toss--Relay} pairs a thrower with the net catcher and requires dialogue
about qualitative rendezvous, orientation, readiness, and observed catch
outcome.  \textbf{Clear--Toss--Deliver} is held out as a compositional test in
which the skills are individually familiar but their adaptive cooperation is
not.  Additional push-or-pick and rearrange-access instances vary which action
is cheapest without adding a new communication or belief mechanism.

These labels specify goals and available embodiments, not required action
sequences.  An episode may admit multiple valid strategies: clear a bridge and
load a cart, push a target into reach, or avoid clearance by coordinating a
toss.  Paired cost interventions vary starts, fuel, and action reliability to
test whether agents select different execution traces from the same scene.

\textbf{Current implementation status.}  The PyBullet stack currently
implements all three role interfaces, separate overhead and egocentric RGB
cameras, collision-checked mobile navigation, articulated picking, physical
basket loading, and carrier delivery of 12-mm particles.  A deterministic
functionality rollout completes two full pick--load--deliver cycles while
logging natural-language-only inter-agent messages, and 59 environment and
controller tests pass. This rollout establishes physical functionality. The
policy input, execution, and messaging boundaries above are requirements for
the planned TRBC evaluation.  Before running the frozen matrix, we will add the
procedural disturbance, independently randomized landmark attributes,
special-target identification, movable-obstacle pushing, physical release and
net catching, both carrier embodiments, and the TRBC belief and dialogue
interfaces.  Each new skill will first pass isolated contact tests and a
privileged solvability rollout before entering autonomous evaluation.

The main benchmark contains four nested suites.  \textbf{Unique-reference}
scenes contain one landmark matching each salient description.
\textbf{Ambiguous-reference} scenes repeat a color or category and require a
compound relation or clarification.  \textbf{Revision} scenes introduce a
failed grasp or push, missed throw or catch, drop, occluded receiver, stale
message, or blocked approach after a reasonable initial belief has formed.
\textbf{Held-out composition} scenes combine unseen layouts, landmark and
affordance permutations, object distributions, receiver embodiments,
capability allocations, and non-contact moving distractors.  The target remains
visually identifiable in the observer image but need not be visible to any
local robot.  Test-time beliefs reset completely, internal identifiers are
reminted, and no episode memory crosses the split.

\begin{table}[t]
\centering
\caption{Controlled evaluation axes.  Factors are crossed with matched random
seeds rather than changed one at a time only.}
\label{tab:eval-axes}
\small
\begin{tabular}{@{}p{0.28\columnwidth}p{\dimexpr0.72\columnwidth-2\tabcolsep\relax}@{}}
\toprule
Axis & Levels \\
\midrule
Spatial shift & development, held-out layout, held-out landmark composition \\
Reference ambiguity & unique, repeated attribute, compound relation, frame ambiguity \\
Visibility & observer-only target, intermittent local glimpse, target visible to all \\
Team composition & pick--cart, push--pick--cart, throw--catch, held-out composition \\
Physical evidence & failed grasp/push, missed throw/catch, drop, empty receiver, blocked route \\
Affordance shift & movable/immovable obstacle, mass, friction, restitution \\
Resource shift & starts, fuel, movement/skill cost, communication/time budget \\
Disturbance & frozen perturbators, moving non-contact distractors, noisy RGB \\
Communication & budget and cost sweeps; delayed, stale, or paraphrased messages \\
\bottomrule
\end{tabular}
\end{table}

\subsection{Research Questions and Experiment Matrix}
\textbf{E1---Belief construction (H1).}  Before the first target-directed
action, we measure each acting manipulator's posterior mass and calibration at
the true target or obstacle region after local observation, after the
observer's description, and after any clarification.  We then measure correct
target/obstacle approach, receiver rendezvous, and grounding success on unique,
ambiguous, and held-out capability scenes.

\textbf{E2---Belief revision (H2).} The primary diagnostic pairs two
replays of a failed handoff. We snapshot the episode after a controlled drop
and before the receiver chooses whether to leave. Both branches start with the
same physics state, receiver view, pre-message belief, and teammate report. We
fix the model proposals, admissible options, and candidate costs. One branch
applies the calibrated message update. The other scores the same options with
the pre-message belief. We measure the change in load/catch belief, the
carrier's first choice (wait, inspect, reposition, or depart), and empty-carrier
departures. Separate closed-loop runs measure recovery latency, repeated
failures, and delivery after missed grasps or pushes, blocked routes, drops,
missed throws or catches, and empty loads.

\textbf{E3---Belief exploitation and communication (H3).}  We sweep message,
query, and model-call budgets and compare task return against communication
cost.  We separately test information directed to the picker, carrier, and
observer as well as pusher, thrower, and catcher roles. Policies choose when to
communicate during these runs. At matched budget caps, we compare the
return--cost frontier against broadcast, always ask, and entropy-triggered
communication. A recipient-conditioning ablation holds the belief updater and
action interface fixed.

\textbf{E4---Generalization and leakage controls.} We permute landmark
attributes independently of location, remint IDs, and permute candidate order.
In a separate replay, we fix images, dialogue, and proposals while swapping or
erasing private beliefs to test whether the selector uses them.  We hold out
layout topologies, landmark/affordance compositions, receiver embodiments, and
the Clear--Toss--Deliver capability composition.  We audit
serialized inputs, delivered messages, and skill bindings for privileged
fields. We also vary evaluator-only metadata with the role-local inputs and
randomness held fixed, and check that the pre-execution decision is invariant.
Coordinate-sharing and centralized-state oracles provide privileged reference
conditions and help identify failures of the shared skills. The main
comparison retains all matched episodes.

\textbf{E5---Emergent cooperation (H4).}  In paired episodes with identical
geometry, we intervene on one factor---agent start, fuel, action cost, obstacle
movability, catcher availability, or observed failure---while withholding any
prescribed plan.  We measure whether TRBC changes handoff mode, acting role,
route, requests, and execution trace when the return-maximizing strategy
changes.  We compare against a fixed skill chain and report strategy diversity,
unnecessary travel/manipulation, recovery switches, and final return.

\subsection{Baselines and Ablations}
The primary baselines are: \textbf{Fixed Skill Chain}, which executes a
task-family-specific role sequence with scripted failure recovery;
\textbf{No-Comm}, in which local robots receive no
observer language; \textbf{Scripted-Broadcast}, which always sends a fixed
target description and status updates; \textbf{Free-Form LLM}, which exposes
the same observation and dialogue interfaces and selects semantic options
without an explicit belief or TRBC belief-update checks;
\textbf{History Policy}, a recurrent/history-only controller trained or
prompted on the same trajectories; \textbf{Belief+Always}, which maintains the
TRBC belief but always broadcasts new evidence; and \textbf{Entropy-Query},
which asks whenever posterior entropy exceeds a tuned threshold. In
\textbf{Always-Ask}, the controller queries an eligible teammate at each
communication opportunity while budget remains. Full TRBC is
compared with \textbf{Coordinate Oracle} and \textbf{Centralized Oracle} upper
bounds that respectively share metric state or plan from complete simulator
state.

Ablations remove one mechanism at a time: spatial belief, behavioral belief,
action-consequence belief, recipient meta-belief, recipient capability/readiness
modeling, calibrated message likelihoods, physical-feedback revision,
clarification queries, or decision-relative communication value.
All non-oracle methods share the observation and message boundaries, skill
bindings, physical execution checks, and resource caps. Free-Form LLM uses
the same backbone and low-level skills as TRBC. Evidence calibration and
belief-dependent scoring are mechanisms under comparison.

\subsection{Metrics and Statistical Protocol}
Task metrics include correct-target delivery, wrong-target grasp, handoff
success, bridge clearance, throw/catch success, empty-carrier departure,
handoff-mode and acting-role choice, completion time, path length, fuel,
physical action count, strategy switches after failure, and discounted return.
Belief metrics include target/obstacle-region negative log likelihood, Brier
score, expected calibration error, relational grounding accuracy,
action-consequence calibration, posterior change after contradiction, and time
to recover.  Resource metrics include addressed messages, clarification
queries, language tokens, model calls, unnecessary travel/manipulation, and
wall-clock decision latency.  We report success--cost Pareto frontiers rather
than a single communication penalty.

Every comparison uses at least 50 matched held-out seeds per primary condition;
we increase to 100 when the paired confidence interval remains wide.  We report
paired bootstrap 95\% confidence intervals, paired success differences, and
effect sizes.  Hyperparameters and thresholds are selected only on development
layouts and then frozen.  Logs retain simulator ground truth for evaluation,
but that process is isolated from prompts, agent memories, and controllers.

\section{Limitations}
The planned study is simulation-only and uses predefined functional roles,
relation vocabulary, and low-level skills; task-level cooperation is emergent,
but the skill library itself is engineered.  Private controllers retain metric
geometry even though agents cannot communicate it.  Throwing, catching, and
movable-obstacle dynamics remain simulator approximations.  Perception,
relation and outcome calibration, and recipient-belief inference are
approximate, and success in a procedural island does not establish equivalence
to human cognition, robustness to open-world dialogue, or real-robot dynamics.
We therefore limit the intended claim to human-compatible, belief-mediated
functional cooperation under controlled partial observability.

\section{Conclusion}
The central scientific test is whether evidence from one robot's view leads
another robot to revise its private belief and whether those private beliefs
let a team form and reorganize a cost-sensitive strategy without a prescribed
role sequence.  TRBC exposes that chain explicitly: visual evidence becomes a
private relational belief, ordinary dialogue is valued for a particular
recipient, and execution outcomes revise both subsequent decisions and the
division of labor.  The frozen experiments will determine whether this
human-inspired structure yields more reliable cooperation across unseen
environments and capability compositions than fluent but belief-free dialogue
or fixed skill chains.

\ifincludeappendix
\appendices
\section{Detailed Belief Updates and Engineering Specification}
This appendix provides the mathematical and software details omitted from the
main paper.  They implement the four system-level definitions in
Eqs.~\eqref{eq:private-belief}--\eqref{eq:unified-selection} but are not
additional scientific claims.

\subsection{Formal Model, Objective, and Recipient Model}
The full task is the finite-horizon communication-enabled Dec-POMDP
$\mathcal M=\langle\mathcal I,\mathcal S,\{\mathcal A_i\}_{i\in\mathcal I},
T,\{\Omega_i\}_{i\in\mathcal I},Z,R,\gamma,H\rangle$.  The latent state
contains object, receiver, landmark, obstacle, bridge, and robot
configurations; target identity; basket/net occupancy and deployment; grasp,
push, release, flight, catch, and contact outcomes; route availability; fuel;
distractor motion; and role progress.  Agent $i$ receives
$o_t^i\sim Z^i(\cdot\mid s_t,a_{t-1})$ and acts from private history
$h_t^i=(o_{0:t}^i,u_{0:t}^i,a_{0:t-1}^i)$.  The team objective used by every
method is
\begin{equation}
\begin{aligned}
J(\pi)=\mathbb E_\pi\Bigg[\sum_{t=0}^{H}\gamma^t
  \big(&r_d n_t^{\mathrm{delivered}}-c_a(a_t)-c_e e_t\\
  &-c_m n_t^{\mathrm{msg}}-c_q n_t^{\mathrm{query}}\\
  &-c_f n_t^{\mathrm{failure}}-c_\tau
  -c_{\mathrm{LM}}n_t^{\mathrm{model}}\big)\Bigg],
\end{aligned}
\label{eq:app-objective}
\end{equation}
where delivery counts only the correct target, $e_t$ is movement/skill fuel,
and $c_\tau$ is elapsed-time cost.  Physical action sets are sampled by role:
observer sensing; picker approach/pick/place; particle or obstacle pushing;
thrower pick/aim/release; cart present/inspect/transport/deliver; and catcher
orient/extend/track/inspect/transport/deliver.  Every role may describe,
report uncertainty, state intent or readiness, request, clarify, correct, or
remain silent.  The shared objective and skill inventories do not specify an
execution order.

For every teammate $j$, agent $i$ maintains the bounded first-order model
\begin{equation}
\mu_t^{i\rightarrow j}(b)=P(b_t^j=b\mid h_t^i),
\label{eq:app-meta-belief}
\end{equation}
represented by $K$ role-conditioned belief particles.  Public actions,
delivered utterances, and visible outcomes propagate or reweight particles;
unobserved private evidence is marginalized.  Only evidence that $i$ knows was
delivered to $j$ may update this model.  TRBC does not recursively model what
$j$ believes about $i$.

\subsection{Prediction, Local Evidence, and Message Evidence}\label{sec:app-updates}
Let $b_{t-1,+}^i$ be the previous posterior.  Prediction and one local update
are
\begin{align}
\bar b_t^i(s,z^{-i})
&=\sum_{s',z'^{-i}}T^i(s,z^{-i}\mid s',z'^{-i},a_{t-1})\\[-2pt]
&\quad\times b_{t-1,+}^i(s',z'^{-i}),\\
b_{t,\mathrm{local}}^i(s,z^{-i})
&\propto Z^i(o_t^i\mid s,z^{-i})\bar b_t^i(s,z^{-i}).
\label{eq:app-local-update}
\end{align}
For a relation $\rho$ between referent $r$ and landmark $\ell$, let
$\psi_\rho(x_r,x_\ell;f)$ be a soft compatibility function under frame $f$.
The recipient likelihood marginalizes landmark association $d$ and frame:
\begin{equation}
\begin{aligned}
L_m^i(x_r)&=\sum_{d\in\mathcal D^i(\ell)}\sum_{f\in\mathcal F^i(F_m)}
P(d,f\mid G_t^i,m)\\
&\quad\times\sum_{x_\ell}\psi_\rho(x_r,x_\ell;f)\,b_{t,\mathrm{local}}^i(x_\ell\mid d).
\end{aligned}
\label{eq:app-relation-likelihood}
\end{equation}
Reliability $\kappa_m$ is calibrated by sender, message family, and confidence
bin on development scenes.  Dependence weight $\alpha_m$ equals one for
independent evidence, is reduced for correlated same-source evidence, and is
zero for duplicates.  For new spatial and behavioral message sets
$\mathcal N_{t,\mathrm{sp}}^i$ and $\mathcal N_{t,\mathrm{beh}}^i$,
\begin{align}
b_{t,+,\mathrm{sp}}^i(x)
&\propto b_{t,\mathrm{local},\mathrm{sp}}^i(x)
\prod_{m\in\mathcal N_{t,\mathrm{sp}}^i}\widetilde L_m^i(x),\\
b_{t,+,\mathrm{beh}}^i(z^{-i})
&\propto b_{t,\mathrm{local},\mathrm{beh}}^i(z^{-i})
\prod_{m\in\mathcal N_{t,\mathrm{beh}}^i}P(y_m\mid z^{-i}).
\label{eq:app-message-products}
\end{align}
For an executed option $a_t^i$, let $\xi$ denote its context-conditioned
outcome state and let $y_t^i$ be the observed physical result.  The
action-consequence belief is revised by intervention-conditioned evidence:
\begin{equation}
b_{t,+,\mathrm{cau}}^i(\xi)
\propto P(y_t^i\mid \xi,a_t^i)\,
b_{t,\mathrm{local},\mathrm{cau}}^i(\xi).
\label{eq:app-causal-update}
\end{equation}
Support identifiers are appended atomically after these updates so one
observation cannot be consumed twice.  Ambiguous references remain distinct
posterior modes.

\subsection{Candidate-Specific Return Approximations}\label{sec:app-valuation}
For a closed-loop physical option $a$, TRBC uses the role-specific belief and a
one-step macro-action model:
\begin{equation}
\widehat Q_{\mathrm{phys}}^i(b^i,a)
=\mathbb E_{s\sim b^i}\!\left[r(s,a)+\gamma
\widehat V^i(T_a(b^i,s))\right].
\label{eq:app-physical-value}
\end{equation}
For an addressed utterance $u:i\rightarrow j$, recipient particles
$b^{j,k}\sim\mu^{i\rightarrow j}$ approximate how the message changes the
recipient's next decision:
\begin{align}
\widehat Q_{\mathrm{msg}}^i(u)
&=-C_m(u)+\frac{\gamma}{K}\sum_{k=1}^{K}\\[-2pt]
&\quad\times Q_{\mathrm{team}}^i\!\left(b^i,
\widehat\pi_j(\tau_j(b^{j,k},m^{j,k}(u)))\right),\\
\widehat{\mathrm{VoC}}^i(u)
&=\widehat Q_{\mathrm{msg}}^i(u)-\widehat Q_{\mathrm{silent}}^i(j).
\label{eq:app-voc}
\end{align}
For query $q:i\rightarrow j$ with admissible replies $\mathcal Y_q$ including
unknown,
\begin{align}
\widehat Q_{\mathrm{query}}^i(q)
&=-C_q(q)+\gamma\sum_{y\in\mathcal Y_q}P(y\mid b^i,q)\\[-2pt]
&\quad\times\max_a \widehat Q_{\mathrm{phys}}^i(\tau_i(b^i,q,y),a),\\
\widehat{\mathrm{VoI}}^i(q)
&=\widehat Q_{\mathrm{query}}^i(q)
-\max_a\widehat Q_{\mathrm{phys}}^i(b^i,a).
\label{eq:app-voi}
\end{align}
Physical, message, query, and silence scores are normalized to the same
expected-return convention before applying Eq.~\eqref{eq:unified-selection}.

\subsection{Internal Records and Evidence Ledger}\label{sec:app-records}
The private implementation contract uses five versioned record families.
\texttt{RoleMemory} stores agent role and embodiment, private skill inventory,
scene graph, spatial, behavioral, and action-consequence beliefs, recipient
particles, evidence ledger, and movement/fuel, action, dialogue, and model-call
budgets.
\texttt{SpatialToken} stores the recipient's private grounding of an utterance:
sender, recipient, message type, referent, landmark, qualitative relation,
reference frame, task predicate, confidence bin, support identifiers,
timestamp, and expiry.  \texttt{Proposal} records a physical option or ordinary
utterance plus proposer and support.  \texttt{VerifiedCandidate} records the
verifier decision, grounded option, precondition snapshot, utility components,
and resource cost.  \texttt{TrajectoryEvent} stores memory hashes, inbound
evidence, proposals, verifier decisions, selected candidate, outcome, cost, and
consumed identifiers.  Records serialize under the closed
\texttt{trbc.v1} schema for implementation and replay only.  The inter-agent
channel carries a natural-language rendering of a verified proposition.
The receiver keys exact replays by consumed delivery event. Related reports
with uncertain independence receive the dependence discount. Replay and
evaluator logs stay outside the acting policy's inputs.

\subsection{Full Engineering Loop and Verifier}\label{sec:app-verifier}
\begin{algorithm}[t]
\caption{Full asynchronous TRBC engineering loop}
\label{alg:trbc-full}
\begin{algorithmic}[1]
\Require role $i$, predicted prior $\bar b_0^i$, memory, schemas, budgets
\While{task is not terminal and a decision budget remains}
  \State Receive one local observation, public events, and addressed utterances
  \State Update the private scene graph from the local observation
  \State Form $b_{t,\mathrm{local}}^i$ from the predicted prior exactly once
  \State Ground utterances against the recipient graph and current belief
  \State Reject malformed, unsupported, consumed, or expired interpretations
  \State Apply Eq.~\eqref{eq:calibrated-message} to each new evidence item once
  \State Update recipient particles from known delivered evidence and public events
  \State Commit posterior, graph, ledger, recipient models, and budgets to memory
  \State Generate a complete deterministic set of physical and silence candidates
  \State Add supported message/query candidates and model semantic proposals
  \State Verify schema, referents, support, capability, preconditions, and budget
  \State Score accepted candidates with Eq.~\eqref{eq:unified-selection}
  \State Execute the fixed-tie argmax or transmit its addressed utterance
  \State Observe outcome, debit resources, and append one trajectory event
  \State Predict $\bar b_{t+1}^i$ from the posterior, action, and outcome
\EndWhile
\end{algorithmic}
\end{algorithm}
The verifier never accepts a model-generated posterior, coordinate, controller
handle, unsupported referent, or motor command.  Provider failure leaves the
deterministic candidate generator and belief-space controller operational.

\subsection{Frozen Defaults, Information Firewall, and Cost}
Unless ablated, recipient models use $K=16$ particles.  Each decision is capped
at eight physical, eight message, and four query candidates plus silence, with
at most one stateless, tool-free, temperature-zero foundation-model call per
acting role.  Spatial descriptions expire after six decision events and
behavioral status after two.  Independent evidence uses dependence weight one,
correlated same-source evidence uses $\alpha_m=0.5$, and duplicate evidence is
rejected.  Prompts expose confidence bins rather than posterior probabilities.
These values, relation vocabulary, calibration map, prompt templates, and all
resource budgets are selected on development layouts and frozen before test.

All referent and evidence identifiers are reminted each episode and never enter
model prompts or utterances.  Scene seeds, stable asset names, simulator state,
utility values, and controller handles are hidden.  The natural-language
channel rejects coordinates, measurements, serialization syntax, identifiers,
and code-like strings.  Candidate order is nonce-permuted rather than sorted by
value, and a separate evaluator process retains simulator ground truth.
Internal records use \texttt{schema\_version=trbc.v1} with enumerated
role/relation/action fields and \texttt{additionalProperties=false}; invalid
provider output never reaches belief update or execution.

For a sparse private graph with $|E^i|$ retained relations, grounding $M$
candidate utterances costs $O(M|E^i|)$.  Candidate valuation is linear in the
bounded candidate count and recipient particle count $K$.  Primary ablations
vary calibration, recipient modeling, physical-feedback revision, communication cost, and lookahead depth while holding the verifier and information firewall fixed.

\fi

\bibliographystyle{plainnat}
\bibliography{references}

\end{document}